\documentclass{article}

\usepackage{arxiv}

\usepackage[utf8]{inputenc} 
\usepackage[T1]{fontenc}    
\usepackage{hyperref}       
\usepackage{url}            
\usepackage{booktabs}       
\usepackage{amsfonts}       
\usepackage{nicefrac}       
\usepackage{microtype}      
\usepackage{lipsum}
\usepackage{graphicx}

\usepackage{amsmath,amssymb,amsfonts}
\usepackage{graphics} 
\usepackage{epsfig} 
\usepackage{times} 
\usepackage{subfig}
\usepackage{tabularx}
\usepackage{booktabs}
\usepackage{multirow}

\usepackage{enumitem}

\DeclareMathOperator*{\argmax}{arg\,max}

\graphicspath{ {./images/} }

\title{PIC4rl-gym: a ROS2 modular framework for Robots Autonomous Navigation with Deep Reinforcement Learning}

\author{
  Mauro Martini \\
  Department of Electronics\\ and Telecommunications (DET)\\
  Politecnico di Torino \\
  Torino, Italy \\
  \texttt{mauro.martini@polito.it} \\
  \And
  Andrea Eirale \\
  Department of Electronics\\ and Telecommunications (DET)\\
  Politecnico di Torino \\
  Torino, Italy \\
  \texttt{andrea.eirale@polito.it} 
  \And
  Simone Cerrato \\
  Department of Electronics\\ and Telecommunications (DET)\\
  Politecnico di Torino \\
  Torino, Italy \\
  \texttt{simone.cerrato@polito.it} \\
  \And
  Marcello Chiaberge \\
  Department of Electronics\\ and Telecommunications (DET)\\
  Politecnico di Torino \\
  Torino, Italy \\
  \texttt{marcello.chiaberge@polito.it} \\
}

\begin{document}
\maketitle
\begin{abstract}
Learning agents can optimize standard autonomous navigation improving flexibility, efficiency, and computational cost of the system by adopting a wide variety of approaches. This work introduces the \textit{PIC4rl-gym}, a fundamental modular framework to enhance navigation and learning research by mixing ROS2 and Gazebo, the standard tools of the robotics community, with Deep Reinforcement Learning (DRL). The paper describes the whole structure of the PIC4rl-gym, which fully integrates DRL agent's training and testing in several indoor and outdoor navigation scenarios and tasks. A modular approach is adopted to easily customize the simulation by selecting new platforms, sensors, or models. We demonstrate the potential of our novel gym by benchmarking the resulting policies, trained for different navigation tasks, with a complete set of metrics.
\end{abstract}


\section{Introduction}\label{sec:intro}
Autonomous navigation algorithms aim at providing mobile robots with efficient planning and control policies to go through cluttered and dynamic environments. Advanced autonomous navigation systems have been explored to improve planners' and controllers' robustness, reliability, and computational efficiency in real-world applications. In the last decade, learning methods have seen a tremendous success among robotics researchers, motivating an increasing collection of innovative works which adopt Deep Reinforcement Learning (DRL) for general autonomous navigation \cite{zhu2021deep}, socially aware path planning \cite{chen2017socially}, and agile aerial vehicles autopilot \cite{song2021autonomous}.
Besides the most common paradigm of sensorimotor agents or local planners, learning agents can be successfully mixed up in alternative ways with the navigation system. Recent works proposed hybrid solutions to optimize classic planners like the Dynamic Window Approach (DWA) \cite{patel2021dwa}. Moreover, \cite{9597689,xiao2022appl} recently showed the effectiveness of the planner's parameters learning approach compared to end-to-end policy learning, resulting in an adaptive optimized planner. 
Despite the impressive research output, many works never reach real robotic applications, and a great amount of them present hard reproducibility limitations. These problems are often tight to the lack of a common tool for robotics to easily develop and compare solutions in the same conditions. Robot Operating System (ROS), recently updated to ROS2 \cite{doi:10.1126/scirobotics.abm6074}, is the standard software exoskeleton for any robotics project. Learning in simulation is certainly the most convenient and safe procedure to train DRL agents for robotics, and Gazebo\footnote{\url{https://gazebosim.org/home}} is the common choice among simulators.
For this reason, we propose the PIC4rl-gym\footnote{Full code accesible at \url{https://github.com/PIC4SeR/PIC4rl_gym}}, an open-source framework in ROS2/Gazebo realized to enhance, simplify and uniform the research on learning-based autonomous navigation, bridging the gap between DRL research and mobile robotics applications. The modular structure of our new gym allows the user to easily adjust the simulation training settings, selecting the desired robotic platform, sensors, neural network architecture, and DRL policy. The importance of this flexibility resides in the fact that autonomous navigation may present infinite different contexts compared to game-like benchmarks where RL algorithms are usually proposed. Moreover, a testing package is also embedded in the gym to encourage the comparison of resulting policies and their transition to real-world applications. It allows to automatically load trained agents and compute navigation metrics for each task of interest in different testing scenarios. We demonstrate the wide potential of the PIC4rl-gym by presenting diverse ablation studies that can be conducted within the framework. These include training and testing navigation policies from scratch and reference to already published works built upon the gym. The resulting benchmarks are reported for each of the considered navigation categories. We want to point out that the PIC4rl-gym does not focus on a specific DRL methodology. It paves the way for developing novel, generic DRL-based navigation solutions, from end-to-end to hybrid approaches.

Therefore, the contributions of this work are:
\begin{itemize}
    \item a modular ROS2/Gazebo framework to develop learning-based navigation solutions for mobile robotic platforms;
    \item a starting collection of training and testing environments, sensors, and neural networks models;
    \item a testing package for trained agents to easily build common benchmarks for each navigation task considered with an established and expandable set of metrics.
\end{itemize}

\section{Related Works}\label{sec:rel_works}
Previous works exist with similar scopes and objectives. 
Multiple versions of a simulation gym for DRL specifically applied to manipulator controls are proposed by \cite{lopez2019gym, nuin2019ros2learn}. \cite{la2022deepsim} is a recent similar work that proposes a gym in ROS, handled with behaviour trees. Regarding this work, we are proposing several advantages with PIC4rl-gym: first, ROS2 presents several benefits and updates and is the actual standard for robotics developers. Second, our parameter-based approach can be better understood by new users for customization, while different behaviour trees may result difficult to be modified. Differently, \cite{perille2020benchmarking} adheres to our idea of a common benchmark for advanced autonomous navigation, proposing interesting metrics to evaluate a difficulty score for each generated environment in Gazebo. Despite the rich collection of Gazebo worlds proposed, this work proposes a dataset of challenging, although not realistic, scenarios. The training framework used to train the adaptive planner with reinforcement and parameter learning approach in \cite{xu2021applr} has not been released. Moreover, it was based on the previous ROS navigation stack and a single platform. 

Nonetheless, disparate research works have already been developed within the PIC4rl-gym framework: end-to-end local planners \cite{martini_mauro_2021_6367976}, position-agnostic vineyard navigation \cite{martini2022position}, and RL-DWA person following \cite{eirale2022monitoring}.

\begin{figure*}[ht]
\centering
\centerline{\includegraphics[width=\textwidth]{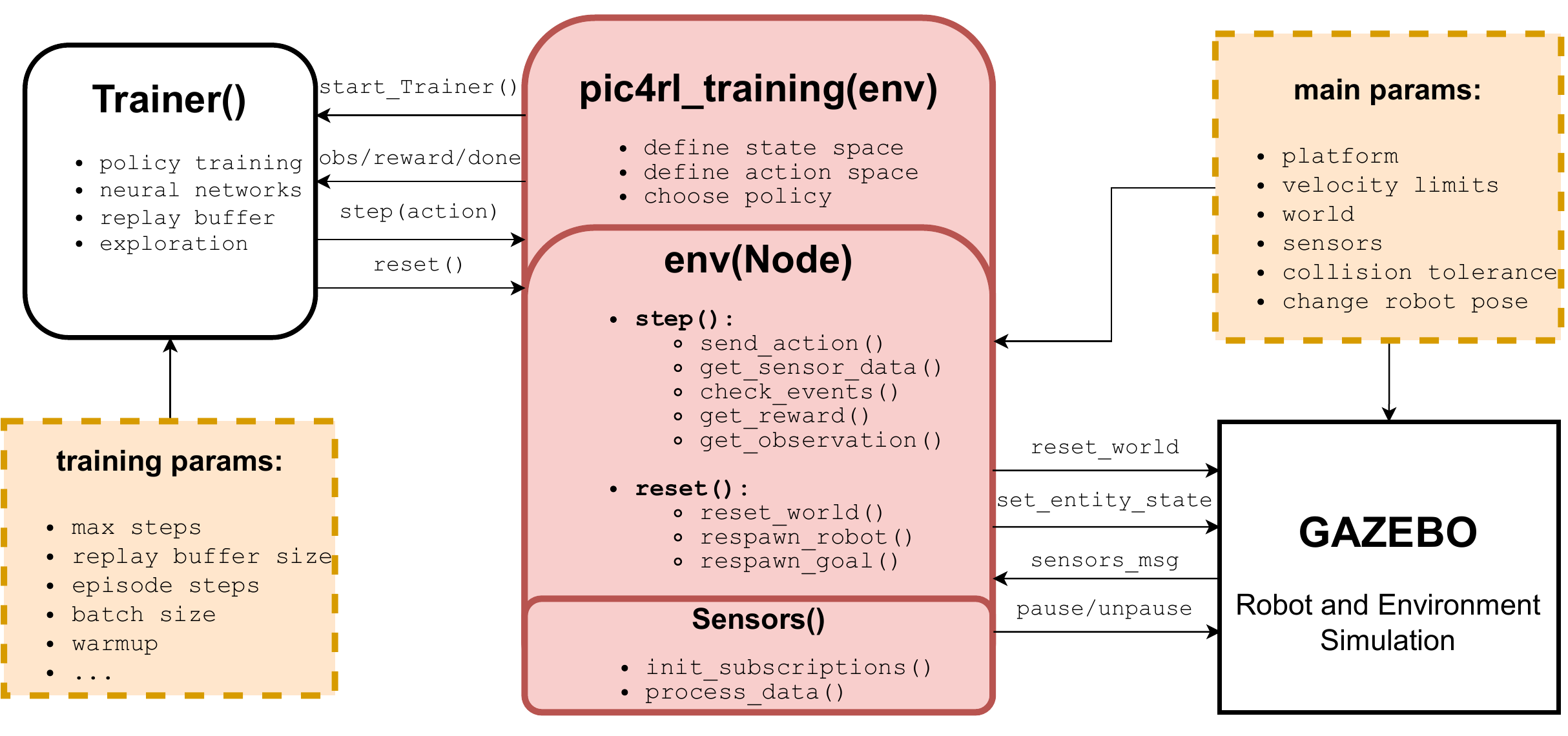}}
\caption{Schematic of ROS nodes and processes composing the PIC4rl gym framework.}
\label{fig:gym_training_scheme}
\end{figure*}

\section{Approach}\label{sec:approach}
In this section, we describe the structure of the PIC4rl-gym and briefly formalize the theoretical framework of Reinforcement Learning for autonomous navigation. The ROS2 nodes architecture to perform parametric training sessions in Gazebo is discussed. Then, the organization of the testing package included in the gym is also presented, introducing the metrics used to evaluate the different robot navigation tasks considered so far.

\subsection{Deep Reinforcement Learning formulation for navigation}
A typical Reinforcement Learning (RL) framework can be formulated as a Markov Decision Process (MDP) described by the tuple $(\mathcal{S},\mathcal{A}, \mathcal{P}, R, \gamma)$. An agent starts its interaction with the environment in an initial state $s_0$, drawn from a pre-fixed distribution $p(s_0)$ and then cyclically selects an action $\mathbf{a_t} \in \mathcal{A}$ from a generic state $\mathbf{s_t} \in \mathcal{S}$ to move into a new state $\mathbf{s_{t+1}}$ with the transition probability $\mathcal{P(\mathbf{s_{t+1}}|\mathbf{s_t},\mathbf{a_t})}$, receiving a reward $r_t = R(\mathbf{s_t},\mathbf{a_t})$.

In reinforcement learning, a parametric policy $\pi_\theta$ describes the agent behavior. In the context of autonomous navigation, we usually model the MDP with an episodic structure with maximum time steps $T$. Hence, the agent's policy is trained to maximize the cumulative expected reward $\mathbb{E}_{\tau\sim\pi} \sum_{t=0}^{T} \gamma^t r_t$ over each episode, where $\gamma \in [0,1)$ is the discount factor. More in detail, we aim at obtaining the optimal policy $\pi^*_\theta$ with parameters $\mathbf{\theta}$ through the maximization of the discounted term:
\begin{equation}
    \pi^*_\theta = \argmax_{\pi} \mathbb{E}_{\tau\sim\pi} \displaystyle \sum_{t=0}^{T} \gamma^t r_t
\end{equation}
which can present alternative expressions according to the specific deterministic or stochastic policy adopted. As mentioned in Section \ref{sec:intro}, the agent's policy can play a wide variety of roles within a navigation framework, depending on the task and the methodological approach. 

\subsection{PIC4rl-gym training framework}
The PIC4rl-gym is designed to provide robotics developers with an easy tool to start custom DRL training sessions in simulation with minimum action on the code. The gym focuses on autonomous navigation tasks, which can be approached with novel solutions based on learning agents or hybrid classic and learning-based navigation components. To this end, we design an extremely modular framework, leveraging ROS parameters for fast simulation tuning. Indeed, two sets of parameters are used to manipulate both the simulation settings and the training details. 
The overall scheme of the PIC4rl-gym framework is shown in Figure \ref{fig:gym_training_scheme}, depicting the organization of ROS nodes and other elements composing the complete system. Arrows indicate how they communicate with the Gazebo simulation environment and the policy Trainer class.

\begin{description}[leftmargin=0pt]
\item[PIC4rl training and environment]
The core section of the gym consists of two elements: a training interface \textit{pic4rl\_training} which bridges the ROS system with the Trainer, and the environment \textit{pic4rl\_environment}. The overall system has been optimized such that training and environment modules are condensed in a single ROS2 node execution, avoiding delays derived from massive data exchange between different nodes. Hence, a class inheritance-based structure is chosen: \textit{training(env(Node))}. The environment reflects the typical design of a DRL gym, presenting two main methods to be called by the Trainer loop: \textit{step()} to get observation and reward data, and \textit{reset()} to restart a new episode, as shown in the schematic in Figure \ref{fig:gym_training_scheme}. Each of the fundamental sub-methods in \textit{step()} defines the specific navigation task. Indeed, different navigation tasks can be tackled by defining a clean environment with appropriate methods to send the predicted action, calling the correct sensor data to process, defining the end-of-episode state condition, and designing the agent's reward function and observation. The associated \textit{pic4rl\_training} will therefore instantiate the environment and the desired training policy, starting the Trainer loop.
ROS2 parameters allow the user to easily customize the training in simulation as requested by the navigation task. For example, it is possible to set episode duration and the number of initial episodes, regulate the exploration, and change the robot's starting pose. This ensemble of practical settings strongly affects the success of the DRL training, being autonomous navigation a much more complex task to learn compared to standard RL benchmarks. Changing the robot pose along the simulation is a fundamental feature for the generalization properties of the agent. In such a way, it is possible to augment the variety of states experienced by the agents without drastically increasing the computational cost of the simulation training. 

\begin{figure}[ht]
\centering
\centerline{\includegraphics[width=0.9\textwidth]{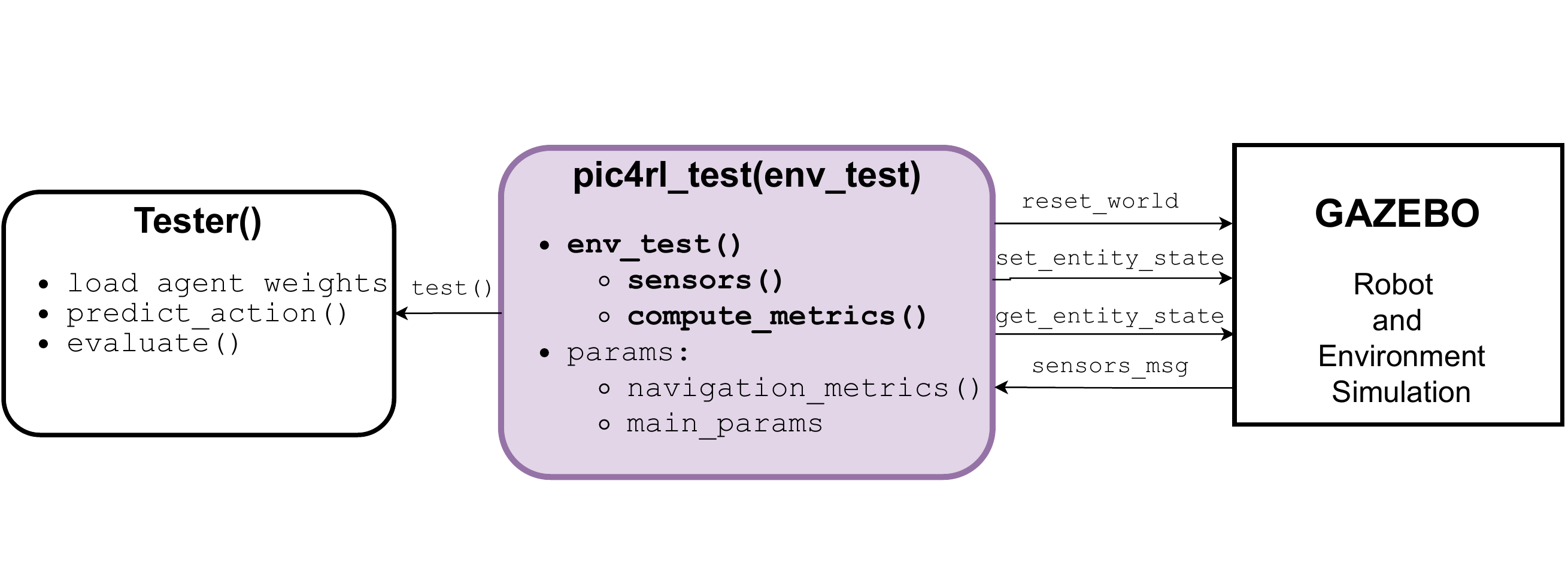}}
\caption{Schematic description of the testing package.}
\label{fig:test_scheme}
\end{figure}

\item[Gazebo simulation] 
It is possible to launch a simulation in Gazebo, choosing the desired world, robotic platform, sensor data, and velocity limits in the main parameters file. In the actual version of the gym, we include standards packages to spawn differential drive platforms such as Jackal\footnote{https://clearpathrobotics.com/jackal-small-unmanned-ground-vehicle/} and Husky\footnote{https://clearpathrobotics.com/husky-unmanned-ground-vehicle-robot/} UGV from ClearPath Robotics or TurtleBot\footnote{https://www.turtlebot.com/}, together with omnidirectional custom platforms \cite{eirale2022marvin}. The structure of the simulation package, which handles the Gazebo virtual environment,  grants to easily add new platforms and sensor plugins. Moreover, the gym automatically manages a precise detection of collisions with obstacles by specifying the shape and desired tolerance for collisions in the parameters. Whenever a \textit{reset()} is performed to start a new episode, the environment calls the \textit{reset\_world} service of Gazebo to reset the simulation to the original state. Then, the gazebo-ros \textit{set\_entity\_state} service is called to re-spawn the robot and the visual model of the goal in a new starting pose. When requested by the task, \textit{pause/unpause} services can be used at each training step to stop the simulation while computing a new action and guarantee the MDP property of the agent-environment interaction loop.

\item[Sensors]
A particular focus is devoted to developing a modular framework for sensors. No available tools offer the possibility to choose or combine data deriving from different sensors for navigation policies with DRL. In PIC4rl-gym, it is possible to specify the desired sensors and the associated properties in the parameters: the number of range points and max distance for 2D LiDAR, resolution and max depth for RGB-D images. The \textit{Sensors()} class will take care of subscribing to the desired topics and process sensor data at each environment step call. Nonetheless, according to the sensor data shape, a set of suitable neural network architectures is already available to be selected in the pic4rl-training node. Simple dense networks are used for 2D LiDAR, while Convolutional neural networks with multiple inputs are implemented to organically deal with feature extraction from images (single-channel or 3-channels) and task-dependent state information such as goal location or robot pose.

\item[Trainer]
The gym's DRL training section is developed by customizing and integrating the TF2RL library \footnote{https://github.com/keiohta/tf2rl} in a ROS2 system.
The \textit{Trainer()} class takes care of interacting with the ROS environment at each step and training the RL policy. Real-world autonomous navigation usually requires agents with continuous action space. According to this, the Trainer focuses on state-of-the-art policy training algorithms which respect this condition. Among them, we have Deep Deterministic Policy Gradients (DDPG) \cite{lillicrap2015continuous}, Twin Delayed DDPG (TD3) \cite{fujimoto2018addressing}, and Soft Actor-Critic (SAC) \cite{haarnoja2018soft}. The TF2RL library also includes diverse replay buffer options, from basic to prioritized and N-step experience replay. The original implementation of the Trainer has been adapted to our gym case integrating it into the modular framework where different network architectures can be selected according to state shape and task and automatically tune the action output shape. Nonetheless, the library only provided an initial warm-up phase for experience collection with a random policy. An $\epsilon$-greedy exploration policy with exponential decay $\gamma_\epsilon$ has also been included to guarantee a continuous exploration rate during the entire training.

\subsection{Testing package}
A testing package has also been developed to easily evaluate trained agents and compute navigation metrics in different testing scenarios. The structure of the \textit{Tester()} reflects the one used for training, only performing policy evaluation. Necessary data are collected from the simulation thanks to ros topics and the Gazebo service \textit{get\_entity\_state}, which allows the robot to store the full traveled path, which can be used to compute metrics for trajectory comparison.
The scheme of the PIC4rl-gym testing package is shown in Figure \ref{fig:test_scheme}.
New metrics can be easily added to the testing framework and selected from the parameters configuration interface.

\end{description}

\section{Experiments and Benchmarks}\label{sec:experiments}
In this section, some representative experimental sessions are described to demonstrate the effectiveness of the PIC4rl-gym on a wide variety of potential applications. Nonetheless, this work aims to introduce the framework without focusing on a specific approach. We report three different case studies, a representative one conducted from scratch, the others referring to already published works realized with the gym.

\begin{figure}[ht]
\centering
\centerline{\includegraphics[width=0.55\textwidth]{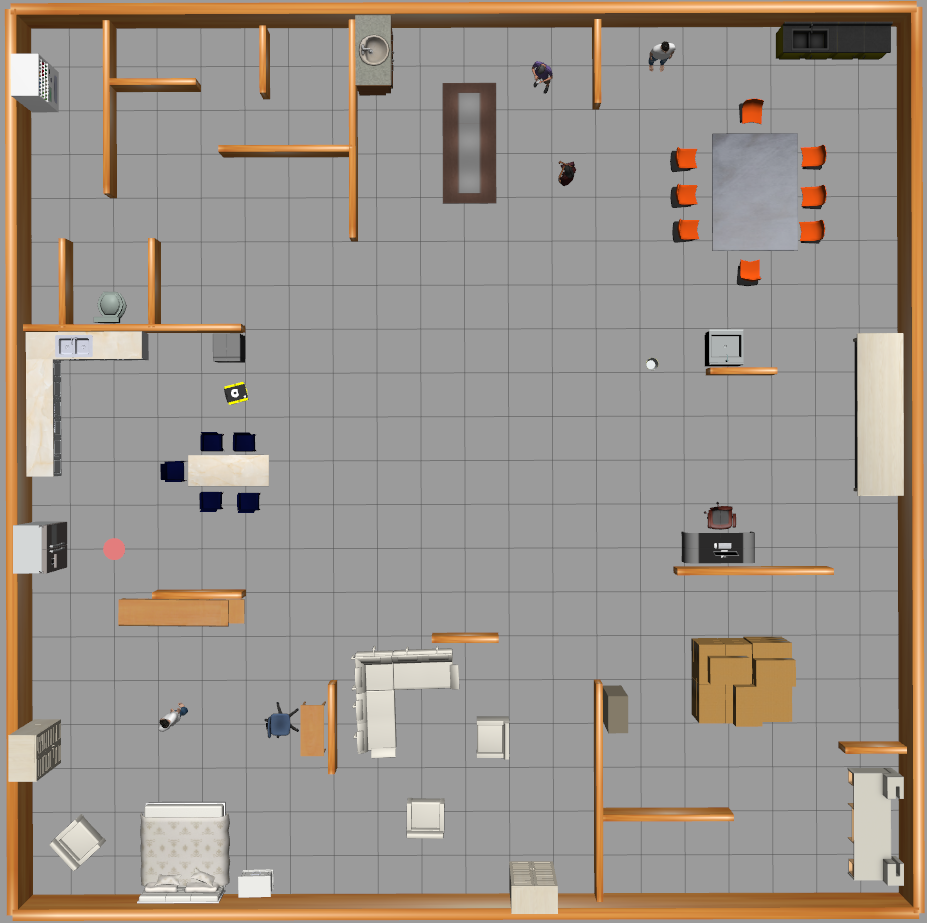}}
\caption{Example of indoor environment in Gazebo where navigation policies can be trained or tested, accessing different areas with realistic obstacles and narrow passages. A Jackal UGV has been spawned to reach the goal (red circular target).}
\label{fig:simulation}
\end{figure}

\subsection{End-to-end policy comparison}
\begin{figure}[ht]
\centering
\centerline{\includegraphics[width=0.9\textwidth]{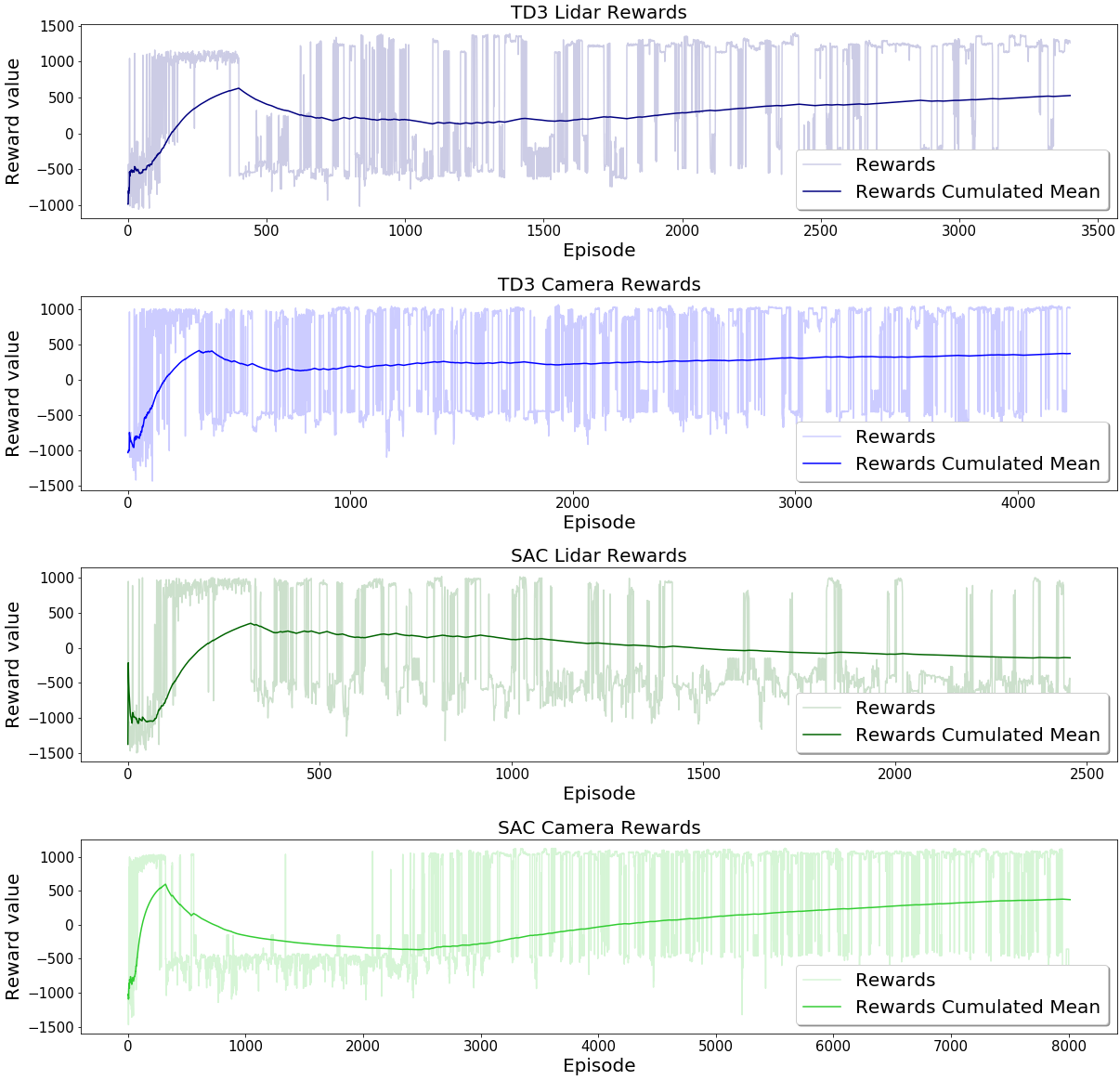}}
\caption{End-to-end policy comparison: reward signals obtained during training the agents. Average reward trend in bold.}
\label{fig:reward_results}
\end{figure}

The classic point-to-point navigation is tackled here with a basic end-to-end policy learning approach, training sensorimotor agents that directly map different sensor data to robot velocity commands [$v,\omega$] for mapless navigation. We conducted an ablation study by training and testing four different policies, changing sensors observation and DRL algorithms. A TD3 and a SAC dense agents are trained with 2D LiDAR ranges as observations, and a TD3 and a SAC convolutional agents are trained with depth images as input. Figure \ref{fig:simulation} shows the indoor world used to train the agents. In this case, a Jackal UGV is used.
Alternative ablation studies can be conducted by varying the velocity ranges, the platform, the reward function, and all the training settings.
Figure \ref{fig:reward_results} shows the reward signal trends obtained during the training. The agents receive a reward $r_g=1000$ if the goal is reached, $r_c=-150$ for collisions, and a dense reward proportional to distance reduction from goal $r_d=(d_{t-1}-d_t)$. The first 300 episodes are conducted in the central area of the world with random goals. All the agents are perfectly able to learn this behaviour, as confirmed by the growing reward trend. Then, the robot is positioned among realistic obstacles, and the agents should adapt to this complex condition. 

The metrics used to evaluate the point-to-point navigation are:
\begin{itemize}
    \item number of successes;
    \item clearance time, cumulative heading average, total path distance, distance/path ratio;
    \item mean and standard deviation of linear and angular velocities, max and mean linear/yaw acceleration; 
    \item min and mean distance from obstacles;
\end{itemize}

Table \ref{tab:navigation-metrics} reports the main results obtained from the testing stage of the study, averaging metrics over five different testing episodes with five different \textit{start-goal} couples assigned around the indoor world.

\begin{table*}[ht]
\centering
\caption{End-to-end policy comparison: average navigation metrics obtained testing the agents on 5 different episodes.}
\label{tab:navigation-metrics}
\resizebox{\textwidth}{!}{%
\begin{tabular}{@{}cccccccccccc@{}}
\toprule
 &success&time&cum. heading&path&dist/path& $v_{mean}$& $\omega_{std.dev.}$&max linear&max yaw&min obstacle& mean obstacle\\
 &[n/total]&[s]&[rad]&[m]&[m/m]&[m/s]&[rad/s]&acceleration[m/$s^2$]&acceleration[rad/$s^2$]&distance [m]&distance [m] \\ \midrule
TD3 LiDAR & 5/5 & 24.11 & 0.10 & 5.84 & 0.96 & 0.28 & 0.54 & 3.41 & 18.73 & 0.37 & 3.08 \\
TD3 Depth & 5/5 & 19.29 & 0.11 & 5.77 & 0.94 & 0.31 & 0.43 & 2.60 & 17.23 & 0.53 & 3.65 \\
SAC LiDAR & 5/5 & 29.36 & 0.28 & 6.06 & 0.94 & 0.25 & 0.44 & 2.98 & 16.98 & 0.31 & 2.66 \\
SAC Depth & 5/5 & 16.65 & 0.13 & 5.74 & 0.98 & 0.36 & 0.41 & 2.51 & 16.58 & 0.42 & 2.78 \\ \bottomrule
\end{tabular}%
}
\end{table*}

\subsection{Row-based crops navigation}
A position-agnostic navigation agent for vineyards row-based crops has been trained in the PIC4rl-gym and already presented in detail in the paper \cite{martini2022position}. In this case, a SAC convolutional agent learns how to safely guide a robot through vineyard rows without localization data. Besides velocity comparison, the robot's trajectory is evaluated in terms of difference from the center of the row. Different Gazebo vineyard worlds are available in the gym for testing. Row-based crops navigation is a fundamental and novel benchmark in the precision agriculture domain that the PIC4rl-gym can significantly boost and uniform.

\subsection{Person monitoring}
The PIC4rl-gym has also been used to develop an RL-DWA peculiar person monitoring algorithm presented in \cite{eirale2022monitoring}. In this alternative assistive navigation task, a DRL agent is trained to keep the orientation of an omnidirectional robot towards the person to be monitored, while the robot is moving avoiding obstacles.

\section{Conclusion and Future Works}
In this paper, we presented \textit{PIC4rl-gym}, a novel modular framework in ROS2/Gazebo to train and test Deep Reinforcement Learning agents specifically for mobile robot navigation tasks. The highly modular and adaptable structure of PIC4rl-gym provides robotics researchers with a complete and fast tool for developing cutting-edge DRL-based navigation solutions for various tasks. Our experiments highlight the large set of possible configurations one can explore by combining different platforms, sensors, policies, and neural network models, leading to performance optimization and flexibility. Nonetheless, we also release a testing package and a standard set of testing environments to enhance the practice of comparing proposed solutions on a common research benchmark.
\begin{itemize}
    \item Increase the variety of training and testing scenarios
    \item Expand available sensors (stereo depth camera and Ultrawide-band anchors)
    \item Include other navigation tasks and benchmarks like exploration and social navigation.
\end{itemize}
\bibliographystyle{unsrt}  
\bibliography{biblio}  

\end{document}